\documentclass[10pt,twocolumn,letterpaper]{article}

\usepackage{iccv}
\usepackage{times}
\usepackage{epsfig}
\usepackage{graphicx}
\usepackage{amsmath}
\usepackage{amssymb}
\usepackage{multirow}

\usepackage[pagebackref=true,breaklinks=true,letterpaper=true,colorlinks,bookmarks=false,citecolor=blue,linkcolor=blue]{hyperref}

\iccvfinalcopy 

\def\iccvPaperID{3164} 

\ificcvfinal\fi

\begin{document}

\title{Towards Interpretable Deep Networks for Monocular Depth Estimation}

\author{
Zunzhi You\\
Sun Yat-sen University\\
{\tt\small youzunzhi@gmail.com}

\and
Yi-Hsuan Tsai\\
NEC Laboratories America\\
{\tt\small ytsai@nec-labs.com}
\\
\and
Wei-Chen Chiu\\
National Chiao Tung University\\
{\tt\small walon@cs.nctu.edu.tw}

\and
Guanbin Li\thanks{Corresponding author is Guanbin Li.}\\
Sun Yat-sen University\\
{\tt\small liguanbin@mail.sysu.edu.cn}


}

\maketitle
\ificcvfinal\thispagestyle{empty}\fi

\begin{abstract}

Deep networks for Monocular Depth Estimation (MDE) have achieved promising performance recently and it is of great importance to further understand the interpretability of these networks. Existing methods attempt to provide post-hoc explanations by investigating visual cues, which may not explore the internal representations learned by deep networks. In this paper, we find that some hidden units of the network are selective to certain ranges of depth, and thus such behavior can be served as a way to interpret the internal representations.
Based on our observations, we quantify the interpretability of a deep MDE network by the depth selectivity of its hidden units.
Moreover, we then propose a method to train interpretable MDE deep networks without changing their original architectures, by assigning a depth range for each unit to select. Experimental results demonstrate that our method is able to enhance the interpretability of deep MDE networks by largely improving the depth selectivity of their units, while not harming or even improving the depth estimation accuracy. 
We further provide comprehensive analysis to show the reliability of selective units, the applicability of our method on different layers, models, and datasets, and a demonstration on analysis of model error.
Source code and models are available at \url{https://github.com/youzunzhi/InterpretableMDE}.

\end{abstract}

\section{Introduction}
\label{intro}

Monocular Depth Estimation (MDE) has drawn a lot of attention since it is critical for further applications like 3D scene understanding or autonomous driving, due to the less requirement and cost compared to depth estimation using stereo image pairs. Eigen \etal~\cite{eigen2014depth} first utilize convolutional neural networks to perform MDE; since then numerous approaches based on deep neural networks have been proposed and significantly improve state-of-the-art performance~\cite{fu2018deep, hu2019revisiting, yin2019enforcing, lee2019big}.
However, only few studies focus on the interpretability of these MDE networks~\cite{zhao2020monocular}. Since depth estimation can be closely related to downstream tasks like autonomous driving, the lack of interpretability on MDE models could potentially cause critical consequences.

\begin{figure}[t]
    \centering
	\includegraphics[width=0.45\textwidth]{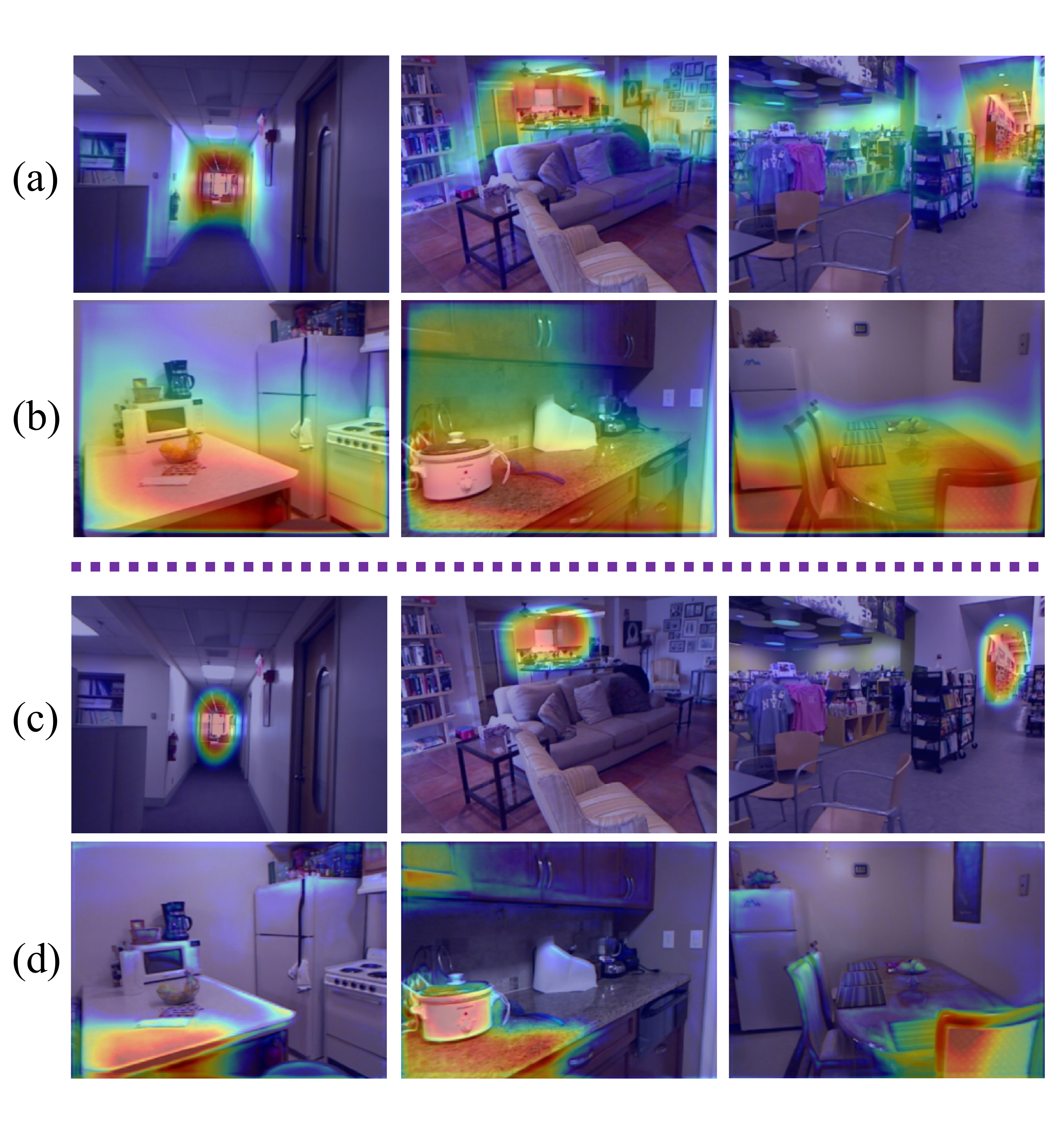}
	\caption{Visualization of feature maps. (a) and (b) refer to the feature map visualization of Unit 5 in layer MFF and Unit 26 in layer D of~\cite{hu2019revisiting} (ResNet-50), respectively. (c) and (d) refer to Unit 63 in layer D and Unit 0 in layer MFF of the interpretable counterpart trained by our method, respectively (best viewed in color). We show that (b) has activations over different depth ranges, while our results in (c) and (d) focus on distant or close depth, which allows more interpretability of the model.
	}
    \label{fig:fmap}
    \vspace{-4mm}
\end{figure}

\begin{figure}[t]
    \centering
	\includegraphics[width=0.45\textwidth]{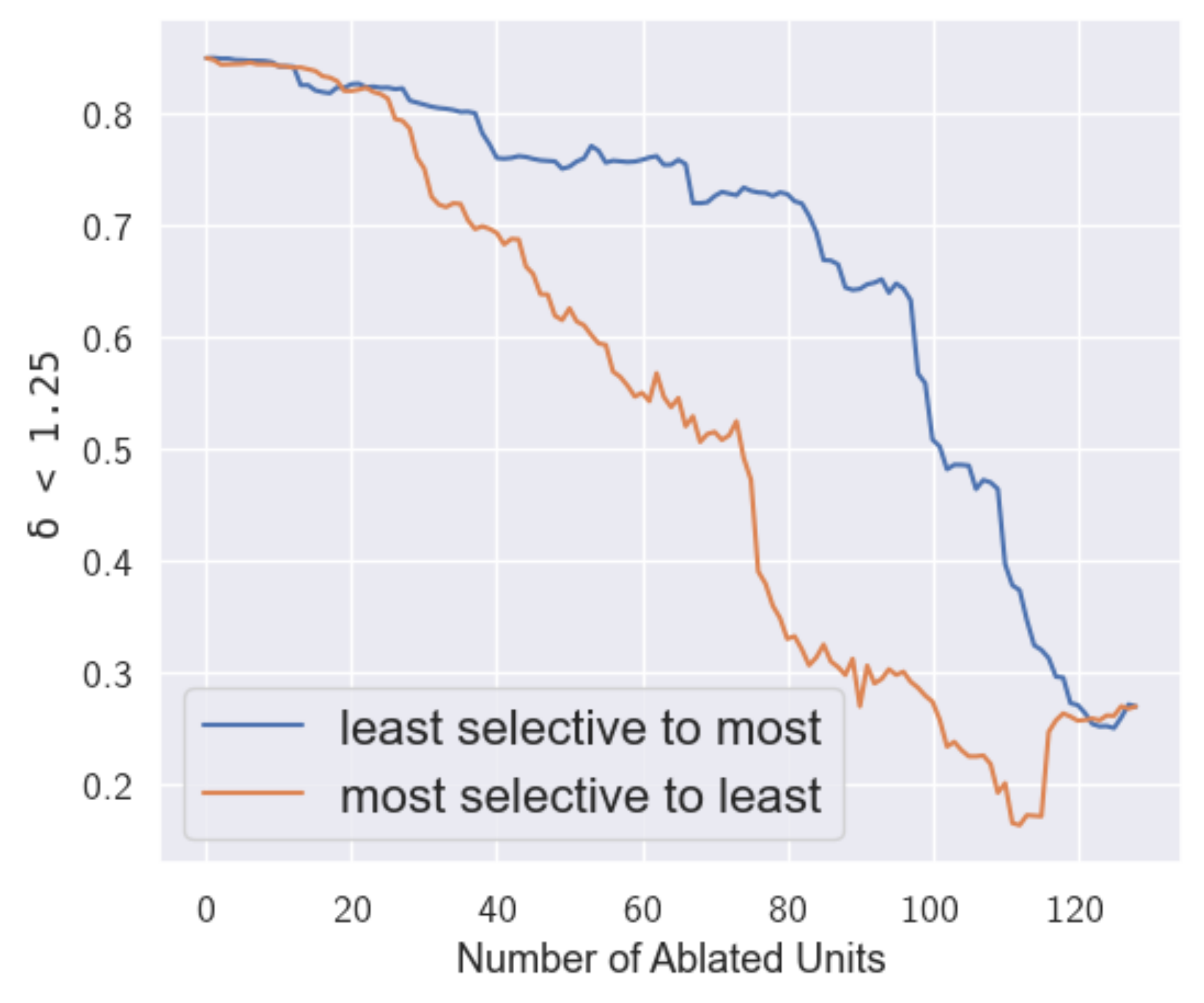}
	\caption{A comparison of the accuracy drop rate when units are ablated successively in different orders. The units are sorted by their depth selectivity and then successively ablated in two reversed order. The accuracy in the y-axis drops faster when units with the higher selectivity are ablated before the less selective ones.}
    \label{fig:ablate}
    \vspace{-4mm}
\end{figure}

In general, understanding deep networks is of great necessity.
Previous works on the interpretability of deep networks for vision mainly focus on image classification~\cite{DBLP:conf/aaai/ZhangCSWZ18, bau2017network} or image generation~\cite{DBLP:conf/iclr/BauZSZTFT19}. On depth estimation, Hu \etal~\cite{hu2019visualization} and Dijk \etal~\cite{dijk2019neural} analyze how deep networks estimate depth from single images by investigating the visual cues in input images, on the level of pixels or semantics, respectively. However, they still treat the networks as black boxes, resulting in less exploration of the internal representations learned by the MDE networks.
In addition, such post-hoc explanations may not present the whole story of interpretable machine learning models as discussed in~\cite{rudin2019stop}.
Although there exists interpretable models for computer vision tasks, such as image classification~\cite{zhang2018interpretable, chen2019looks}, object detection~\cite{Wu2017TowardsIR} or person re-identification~\cite{liao2019interpretable}, these tasks have quite a different characteristics from MDE and are not directly applicable to MDE.
%

\begin{figure}[t]
    \centering
	\includegraphics[width=0.5\textwidth]{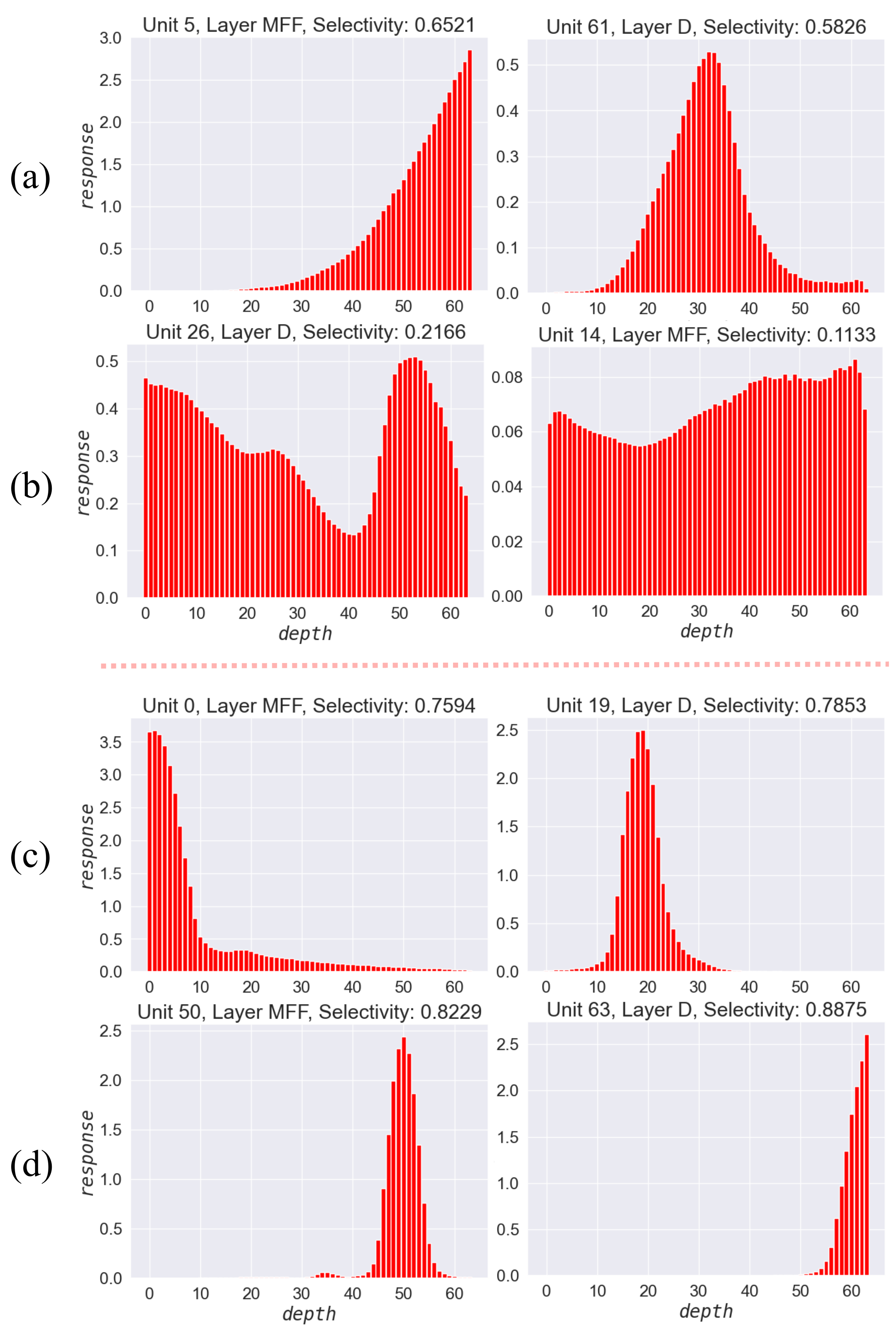}
	\caption{Dissection results on units. (a) and (b) are units of layer D and layer MFF in~\cite{hu2019revisiting} (ResNet-50), where it shows diverse ranges of selectivity. Using our proposed interpretable model, we consistently increase the selectivity over all the units, e.g., (c) and (d), which improves the model interpretability.
	}
    \label{fig:dissection}
    \vspace{-4mm}
\end{figure}

Recently, numerous methods try to discover what neurons in neural networks look for~\cite{olah2017feature, bau2017network, fong2018net2vec, rafegas2020understanding}. It is shown that neuron units generally extract features that can be interpreted as various levels of semantic concept, from textures and patterns to objects and scenes. Moreover, to learn interpretable neural networks, one option is to disentangle the representations learned by internal filters, which makes the filters more specialized~\cite{zhang2018interpretable, liang2020training}. 
Inspired by these works, we observe that in deep MDE networks, some hidden units are selective to some ranges of depth. For example, in Fig.~\ref{fig:fmap}(a), we visualize several feature maps of one unit in a layer of the network from~\cite{hu2019revisiting}. This unit is obviously more activated in the distant regions of the input images. We further dissect the units by collecting their averaged response on depth ranges (see Section~\ref{dissect} later for more details), and Fig.~\ref{fig:dissection}(a) shows that for some units, activations are higher for some certain ranges of depth. 

To quantify this observation, we then compute its depth selectivity for each unit (detailed in Section~\ref{selectivity}). To evaluate the meaningfulness of depth selectivity, we successively ablate units and see how the performance of the network drops accordingly. We first sort the 128 units of the network from~\cite{hu2019revisiting} by their selectivity and then successively ablate units from the most selective unit to the least one, and then do the same thing similarly in the reversed way. In Fig.~\ref{fig:ablate}, the performance of the MDE model drops much quicker when more selective units are ablated earlier than less selective ones. Based on the observations stated above, we argue that for an MDE deep network, a unit is more important when it is more depth selective, and the behavior of its units can be interpreted by telling which ranges of depth activated most by those units. Therefore, the interpretability of a deep network for MDE can be quantified by the depth selectivity of its internal units.

However, in the existing MDE model, despite that some units can be interpreted as being selective for some ranges of depth, most of them have little interpretability.
For example, Fig.~\ref{fig:fmap}(b) and Fig.~\ref{fig:dissection}(b) show feature map visualizations and dissection results of typical units in the network from~\cite{hu2019revisiting}, which have less interpretability.
Therefore, to achieve an MDE model with better interpretability, we propose a simple yet effective interpretable deep network for MDE by maximizing the selectivity of internal units. Our method can be applied to existing deep MDE networks without modifying their original architectures or requiring any additional annotations.
More importantly, we show that it is possible to learn our interpretable model without harming its depth performance, which creates potential discussions in explainable AI along the trade-off between interpretability and model performance~\cite{Rudin2019Why}.
The experimental results show that the our interpretable models achieve competitive or even better performance than the original MDE models, while the interpretability is largely improved.

\vspace{-4mm}
\paragraph{Contributions.} {
To summarize, this work has the following contributions: (1) we quantify the interpretability of deep networks for MDE based on the depth selectivity of models' internal units; (2) we propose a novel method to learn interpretable deep networks for MDE without modifying the original network's architecture or requiring any additional annotations; (3) we empirically show that our method effectively improves the interpretability of deep MDE networks, while not harming or even improving the depth accuracy, and further validate the reliability and applicability of the proposed method.
}

\section{Related Work}
\label{related}

\subsection{Monocular Depth Estimation}
Estimating depth from images is an important problem towards scene understanding, and recently monocular depth estimation has been studied extensively.
Numerous methods based on deep convolutional neural networks have been proposed to achieve better performance on this task, including the usage of geometric constraints, adopting multi-scale network architecture, or sharing features with semantic segmentation~\cite{laina2016deeper, fu2018deep, hu2019revisiting, yin2019enforcing, jiao2018look, lee2019big, zhu2020edge}. Nevertheless, few studies analyze what these deep networks have learned. By modifying input images, Dijk \etal~\cite{dijk2019neural} investigate the visual cues of what a network~\cite{godard2017unsupervised} exploits when predicting the depth. Hu \etal~\cite{hu2019visualization} hypothesize that deep networks can estimate depth from only a selected set of image pixels fairly accurately, and train another network to predict those pixels. Despite that some of their findings are interesting and useful to help understand deep networks for MDE, they still treat the networks as black boxes and their post-hoc explanations do not lead to inherently interpretable models.

\subsection{Post-hoc Explanations for Deep Networks}
Recently, many studies aim to explain deep networks in a post-hoc fashion. Among them, a line of research can be categorized into saliency methods or attribution methods, where the ``important'' pixels are highlighted in input images for networks to give their predictions~\cite{zeiler2014visualizing, selvaraju2017grad, DBLP:conf/icml/SundararajanTY17, DBLP:conf/iclr/KindermansSAMEK18, wang2019learning}. While some recent studies discuss their reliability~\cite{DBLP:series/lncs/KindermansHAASDEK19, sundararajan2017axiomatic, adebayo2018sanity}, these methods are not directly applicable to the task of MDE, since MDE is required to predict a depth value for every pixel, and thus it is not reasonable to use highlighted pixels to attribute the dense prediction of all pixels.

Another group of studies on interpretability of deep neural networks explore the properties or the behavior of single units~\cite{zeiler2014visualizing, DBLP:journals/corr/ZhouKLOT14, bau2017network, olah2017feature, olah2018the, DBLP:conf/iclr/MorcosBRB18, DBLP:conf/iclr/BauZSZTFT19, rafegas2020understanding}, where our work generally falls in this group as we quantify the interpretability of networks for MDE. The fundamental difference between the task of MDE and image classification makes our work distinct from theirs. Moreover, these methods still focus on the explanations of deep networks, instead of designing interpretable models. 

\subsection{Interpretable Deep Networks for Vision}
Instead of providing explanations, some studies attempt to design inherently interpretable models to alleviate the lack of model interpretability in computer vision tasks.
Chen \etal~\cite{chen2019looks} propose an interpretable model for object recognition that finds prototypical parts and reasons from them to make final decisions. Liao \etal~\cite{liao2019interpretable} propose an approach to enhance the interpretability of person re-identification networks by making the matching process of feature maps explicit.
Moreover, other methods that share a similar concept to our method are to learn more specialized filters.
In interpretable CNNs from~\cite{zhang2018interpretable}, each filter represents a specific object part, while a more recent study~\cite{liang2020training} trains interpretable CNNs by alleviating filter-class entanglement, i.e. each filter only responds to one or few classes.
In this paper, our proposed interpretable model focuses on the MDE task by increasing the depth selectivity of units internally in MDE models, which differs from the aforementioned approaches.
\section{Interpretability of Deep Networks for MDE}
\label{interpretability}

In this section, we present how we quantify the interpretability of the units by calculating their depth selectivity with their average response on different ranges of depth.

\subsection{Average Response of Units on Depth}
\label{dissect}

We first dissect a deep network for MDE by collecting the average response of its units on depth. Denote images and the corresponding depth maps in a depth dataset $D$ as $(\mathbf{x}_i, \mathbf{d}_i)\in D$, where $i \in \{1,2,...,N\}$ and $N$ is the number of samples in $D$.
For every internal unit $k$ in a layer $l$ of the deep network, the activation map $A_{l,k}(\mathbf{x}_i)$ is scaled up to the resolution of depth map using bilinear interpolation, denoted as $\tilde{A}_{l,k}(\mathbf{x}_i)$. Depth values in $\mathbf{d}_i$ can be discretized into $N_b$ bins to capture the meaningful depth distribution.
Then, for every discretized depth value $d$ (\ie, the index of a bin) in the discretized depth map $\hat{\mathbf{d}}_i$, we can obtain a binary mask $M_i^{d}$ calculated by $\mathbb{I}(\hat{\mathbf{d}}_i = d)$, where $\mathbb{I}(\cdot)$ is the indicator function. The average response $R_{l,k}^{d}$ of unit $k$ in layer $l$ for depth $d$ is then computed over the entire dataset:

\begin{equation}
\label{equ:response}
\begin{aligned}
R_{l,k}^{d} = \frac{\sum_{i=1}^N S(\tilde{A}_{l,k}(\mathbf{x}_i)\odot M_i^{d})}{\sum_{i=1}^N S(M_i^{d})},
\end{aligned}
\end{equation}
where $S(\cdot)$ sums over all the elements of a matrix and $\odot$ denotes the element-wise multiplication.

\subsection{Depth Selectivity}
\label{selectivity}

Based on the average response, we compare how each unit is activated by different depth ranges and observe that some units are selective to a certain range of depth. Inspired by the commonly-used selectivity index in systems neuroscience~\cite{de1982orientation, britten1992analysis, freedman2006experience}, Morcos \etal~\cite{DBLP:conf/iclr/MorcosBRB18} propose a metric to calculate the class-selectivity of a unit based on its class-conditional average activity, for the task of image classification. Here we adopt this metric to the domain of depth estimation. We define the depth selectivity of a unit as:

\begin{equation}
\label{equ:selectivity}
\begin{aligned}
DS_{l,k} = \frac{|R_{l,k}^{max}|-|\bar{R}_{l,k}^{-max}|}{|R_{l,k}^{max}|+|\bar{R}_{l,k}^{-max}|},
\end{aligned}
\end{equation}
where $|R_{l,k}^{max}|$ is the absolute value of the max response of unit $k$ in layer $l$ over all discretized depth $d$, and $|\bar{R}_{l,k}^{-max}|$ is the average of all the other non-maximum absolute responses. We use the absolute value to make it applicable for units that may have negative output (e.g., units that use ELU~\cite{clevert2015fast} as the activation function). The value of $DS$ is in the range $[0,1]$, and a $DS$ value close to $1$ indicates that corresponding unit is highly selective (e.g., Fig.~\ref{fig:dissection}(c)(d)). 
To give a more concrete idea about this quantity, we calculate its expectation when unit's response is totally randomized (see supplementary material for the derivation).

\begin{equation}
\label{equ:random_baseline}
\begin{aligned}
\mathbb{E}_{|R_{l,k}^{d}|}[DS_{l,k}] = \frac{1}{3}, \quad |R_{l,k}^{d}|\sim U[0,b],
\end{aligned}
\end{equation}
where $b$ is an arbitrary positive number as the upper bound of $|R_{l,k}^{d}|$, in which its value would not affect the outcome of the expectation.  
This expectation can be considered as a random baseline to be further compared with the depth selectivity of actual MDE networks.

\section{Interpretable Deep Networks for MDE}
\label{interpretable}

As motivated previously, here we would like to consider an important problem: \textit{Is it possible to enhance the interpretability of an MDE deep network without modifying its architecture and harming its performance?} 
In this section, we first present a na\"{\i}ve thought (i.e., regularizing selectivity) together with pointing out its potential issue, and then describe our proposed approach (i.e., assigning depth ranges to units).

\subsection{Regularizing Selectivity}
\label{regularize}
As we have the metric of depth selectivity to quantify the interpretability, we in turn aim to enhance the interpretability of an MDE network by increasing its depth selectivity. 
A straightforward approach that first comes to our mind is adding an additional regularization term $\mathcal{L}_{reg}$ to the objective of the MDE model, which encourages the depth selectivity of all the units in layer $l\in L$ to increase:

\begin{equation}
\label{equ:regularize}
\begin{aligned}
\mathcal{L}_{reg} =& - \lambda\sum_{l\in L}\frac{1}{K_l}\sum_{k}DS_{l,k}\\
 =& - \lambda\sum_{l\in L}\frac{1}{K_l}\sum_{k}\frac{|R_{l,k}^{max}|-|\bar{R}_{l,k}^{-max}|}{|R_{l,k}^{max}|+|\bar{R}_{l,k}^{-max}|},
\end{aligned}
\end{equation}
where $K_l$ is the number of units in layer $l$, and $\lambda > 0$ is a hyperparameter to balance between the original depth estimation loss and our regularization term of depth selectivity. 

\begin{figure}[t]
    \centering
	\includegraphics[width=0.45\textwidth]{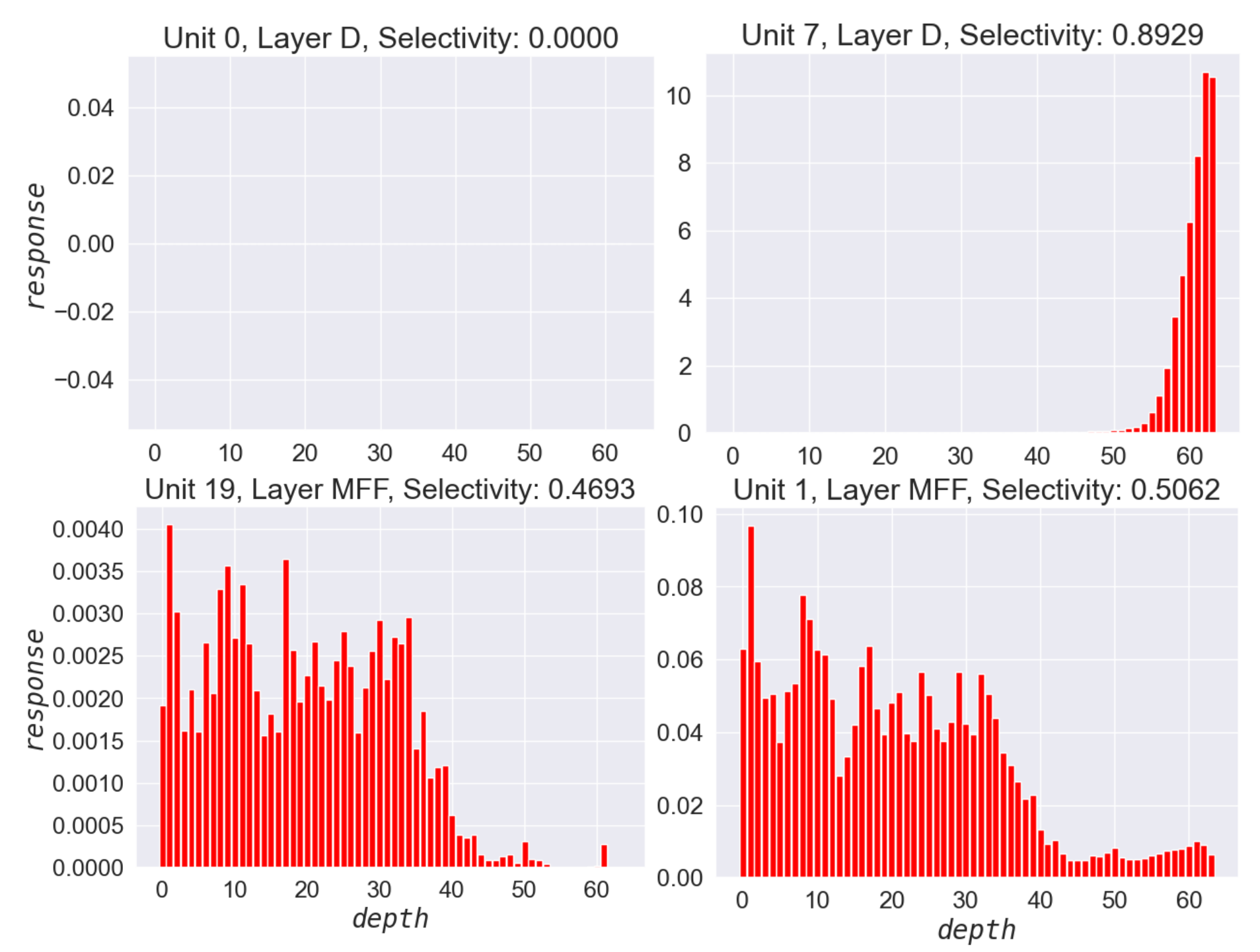}
	\caption{Dissection results of typical units in networks trained by the approach of directly regularizing the depth selectivity via \eqref{equ:regularize}.
	}
    \label{fig:reg}
    \vspace{-2mm}
\end{figure}

However, we experimentally find that such na\"{\i}ve approach leads to unsatisfactory results. 
Fig.~\ref{fig:reg} shows the dissection results of some units in the network trained via regularizing the depth selectivity. 
Despite that some units are still depth selective as we expect, many others are not or collapse (i.e., having no response to any depth values).
This is due to the fact that, during the process of batch-wise optimization, the discretized depths that activate units are mostly different within each batch. Here can be two reasons: (1) at the beginning of training, units are not selective at all, and (2) even if a unit is depth selective, the selected depth could be absent in a batch, and then the unit will be encouraged to activate more on the other depth ranges (e.g., focusing on the range that it activates most in this batch).

\begin{table*}[!t]
	\caption{Depth selectivity and performance of baseline networks for MDE from~\cite{hu2019revisiting} and our interpretable counterparts.}
	\begin{center}
		\begin{tabular}{c|cc|ccc|ccc}
		    Model & Training & Testing & $\delta_{1.25}$ & $\delta_{1.25^2}$ & $\delta_{1.25^3}$ & RMS & REL & log10 \\
		    \hline\hline
		   ~\cite{hu2019revisiting} (ResNet-50) & 0.4617 & 0.4286 & 0.849 & 0.972 & \textbf{0.994} & 0.443 & 0.124 & 0.054 \\
		    Interpretable~\cite{hu2019revisiting} (ResNet-50)  & \textbf{0.8357} & \textbf{0.7529} & \textbf{0.861} & \textbf{0.973} & \textbf{0.994} & \textbf{0.422} & \textbf{0.119} & \textbf{0.051} \\
		    \hline
		   ~\cite{hu2019revisiting} (SENet-154) & 0.4906 & 0.4691 & 0.874 & \textbf{0.979} & \textbf{0.995} & 0.409 & 0.111 & 0.049 \\
		    Interpretable~\cite{hu2019revisiting} (SENet-154) & \textbf{0.8411} & \textbf{0.7693} & \textbf{0.882} & \textbf{0.979} & \textbf{0.995} & \textbf{0.396} & \textbf{0.109} & \textbf{0.047} \\
		\end{tabular}
	\end{center}
	\label{table:result}
\end{table*}

\subsection{Assigning Depth to Units}\label{assign}
In order to tackle the above-mentioned issue happened while regularizing the depth selectivity, we propose a simple yet effective method by assigning each unit a specific depth range for it to select, which is realized by an objective function $\mathcal{L}_{assign}$:
\begin{equation}
\label{equ:assign}
\begin{aligned}
\mathcal{L}_{assign} = - \lambda\sum_{l\in L}\frac{1}{K_l}\sum_{k}\frac{|R_{l,k}^{d_k}|-|\bar{R}_{l,k}^{-d_k}|}{|R_{l,k}^{d_k}|+|\bar{R}_{l,k}^{-d_k}|},
\end{aligned}
\end{equation}
where $d_k$ is the discretized depth being assigned to unit $k$.
As a result, the calculation of selectivity of a unit $k$ is now based on the assigned discretized depth $d_k$, where $|\bar{R}_{l,k}^{-d_k}|$ is the average of all other absolute responses other than $d_k$.
The assignment of depth range to units is based on the following principle:

\begin{equation}
\label{equ:dk}
\begin{aligned}
d_k=\lfloor \frac{k} {{K_l}/{N_b}} \rfloor,
\end{aligned}
\end{equation}
where the number of depth bins $N_b$ is set to $K_l$ if $K_l \leq N_b$, such that every discretized depth $d$ is assigned to at least one unit.
If $d_k$ is absent in a batch, the unit $k$ will be simply disregarded from the computation of $\mathcal{L}_{assign}$.
As a result, this approach does not suffer from problems caused by batch sampling.
Moreover, the interpretability of the deep network is enhanced from another perspective: the behavior of a unit becomes interpretable and predictable as it is now specifically assigned to a particular depth.
Please note that in the following sections and experiments, such proposed approach of assigning depth to units is abbreviated to ``our method'' unless otherwise stated.

\section{Experimental Results}
\label{experiments}

For simplicity, we follow the choice in~\cite{hu2019visualization} and use the network proposed in~\cite{hu2019revisiting} as our target model on the NYUD-V2 dataset~\cite{silberman2012indoor} to show experimental results of our method. We first choose the layer after the multi-scale feature fusion module (referred as layer ``MFF'') and after the decoder module (referred as layer ``D'') in~\cite{hu2019revisiting} to apply our method, as these two layers are closer to the depth output. Nevertheless, we also demonstrate that our approach can be applied to other layers, models, and datasets, and validate the applicability of our method in Section~\ref{more}.
For networks from~\cite{hu2019revisiting}, we consider two variants with different backbones, i.e., ResNet-50~\cite{he2016deep} and SENet-154~\cite{hu2018squeeze}. During training, we follow exactly the same training scheme with the original implementation, including data augmentation, optimizers, total training epochs, etc. 

We set the number of discretized depth bins $N_b$ to 64, since the number of units in most of deep networks for MDE is a power of two, which enables simpler assignment of depth to units. Space-increasing discretization proposed in~\cite{fu2018deep} is adopted to discretize the depth maps, and $\lambda$ in \eqref{equ:assign} is set to $0.1$.

\begin{figure}[!t]
    \centering
	\includegraphics[width=0.45\textwidth]{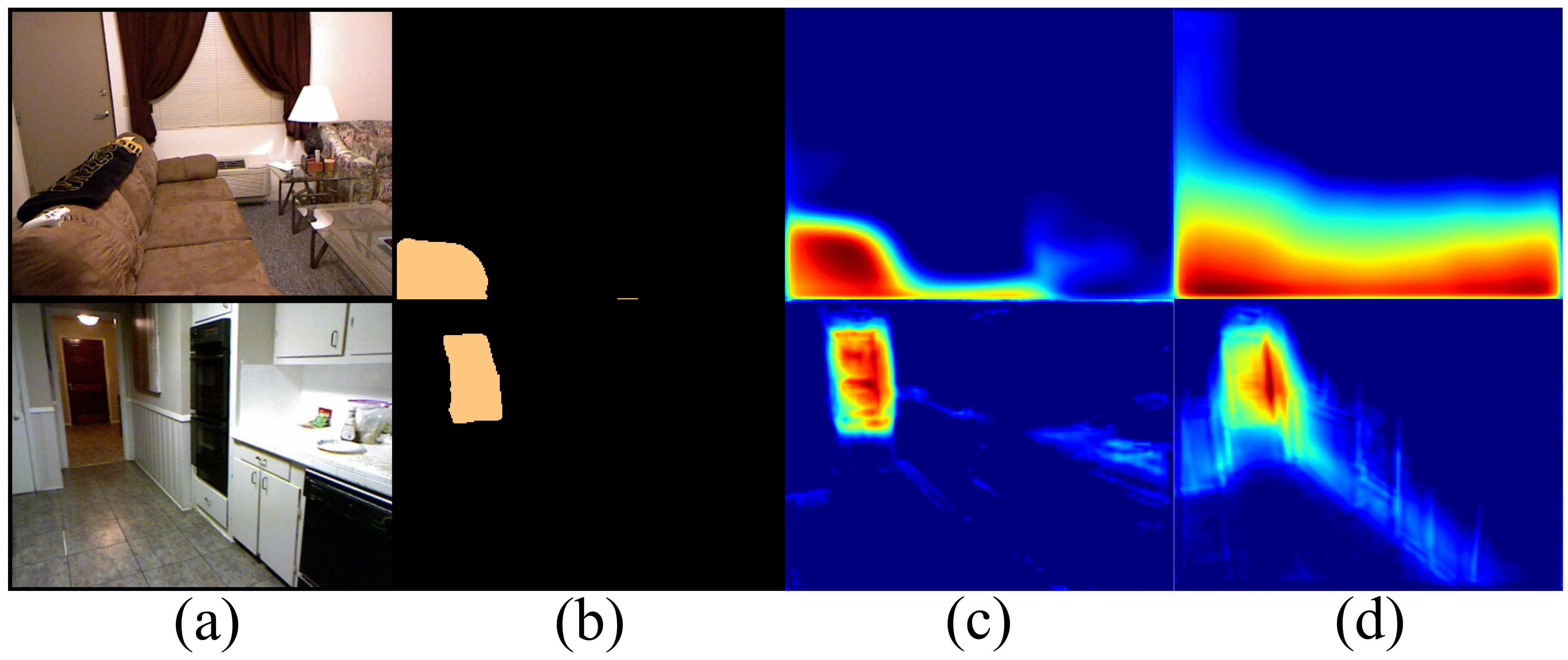}
	\caption{Comparison of the feature maps from our interpretable model and the baseline model~\cite{hu2019revisiting}. (a) Input images. (b) Mask of pixels whose predicted depth is assigned to the corresponding units. (c) Feature maps of our selective units. (d) Feature maps of units in the baseline.
	}
    \label{fig:visual}
    \vspace{-2mm}
\end{figure}

\begin{table*}[!t]
    \caption{Comparison of direct regularizing selectivity (cf. Section~\ref{regularize}) and assigning depth to units (cf. Section~\ref{assign}).}
	\begin{center}
		\begin{tabular}{c|c|cc|ccc|ccc}
		     & & \multicolumn{2}{|c|}{Selectivity $\uparrow$} & \multicolumn{3}{|c|}{Depth Accuracy $\uparrow$} & \multicolumn{3}{c}{Depth Error $\downarrow$} \\
		    Model & Method & Training & Testing & $\delta_{1.25}$ & $\delta_{1.25^2}$ & $\delta_{1.25^3}$ & RMS & REL & log10 \\
		    \hline\hline
		    \multirow{2}{*}{\cite{hu2019revisiting} (ResNet-50)} & $\mathcal{L}_{reg}$ in \eqref{equ:regularize}  & 0.7417 & 0.6039 & 0.857 & \textbf{0.973} & 0.993 & 0.428 & 0.121 & 0.052  \\
                & $\mathcal{L}_{assign}$ in \eqref{equ:assign} & \textbf{0.8357} & \textbf{0.7529} & \textbf{0.861} & \textbf{0.973} & \textbf{0.994} & \textbf{0.422} & \textbf{0.119} & \textbf{0.051} \\
		    \hline
		    \multirow{2}{*}{\cite{hu2019revisiting} (SENet-154)} & $\mathcal{L}_{reg}$ in \eqref{equ:regularize} & 0.7314 & 0.5694 & 0.881 & 0.978 & \textbf{0.995} & 0.399 & \textbf{0.109} & \textbf{0.047} \\
		        & $\mathcal{L}_{assign}$ in \eqref{equ:assign} & \textbf{0.8411} & \textbf{0.7693} & \textbf{0.882} & \textbf{0.979} & \textbf{0.995} & \textbf{0.396} & \textbf{0.109} & \textbf{0.047} \\
		\end{tabular}
	\end{center}
	\label{table:regularizing}
	\vspace{-6mm}
\end{table*}

\subsection{Depth Selectivity and Performance}

\paragraph{Setting and Evaluation Metric.}
First, we conduct experiments to compare the depth selectivity and performance of the baseline models with our interpretable counterparts. We calculate depth selectivity on both training and testing datasets. For depth estimation performance, we follow previous works on MDE to use the following metrics: accuracy under threshold ($\delta_i < 1.25^i, i=1,2,3$), root mean squared error (RMS), mean absolute relative error (REL) and mean $log_{10}$ error (log10). 

\vspace{-4mm}
\paragraph{Main Results.}
It is first observed in Table~\ref{table:result}
that depth selectivity of baseline models is above the random baseline, $1/3$, indicating that MDE deep networks have some level of depth selectivity in their units, while our interpretable models achieve much higher depth selectivity on both training and testing datasets. Fig.~\ref{fig:fmap} (c)(d) also visualize the feature maps of some units in our interpretable networks. Qualitatively, it is shown that these units are more activated on the regions of input images where the depth is assigned to them based on their indices, e.g., distant or close regions. In Fig.~\ref{fig:dissection} (c)(d), we further plot the dissection results of some units in our interpretable networks, showing the selectivity is consistent through the entire dataset. 

\begin{figure}[!t]
    \centering
	\includegraphics[width=0.45\textwidth]{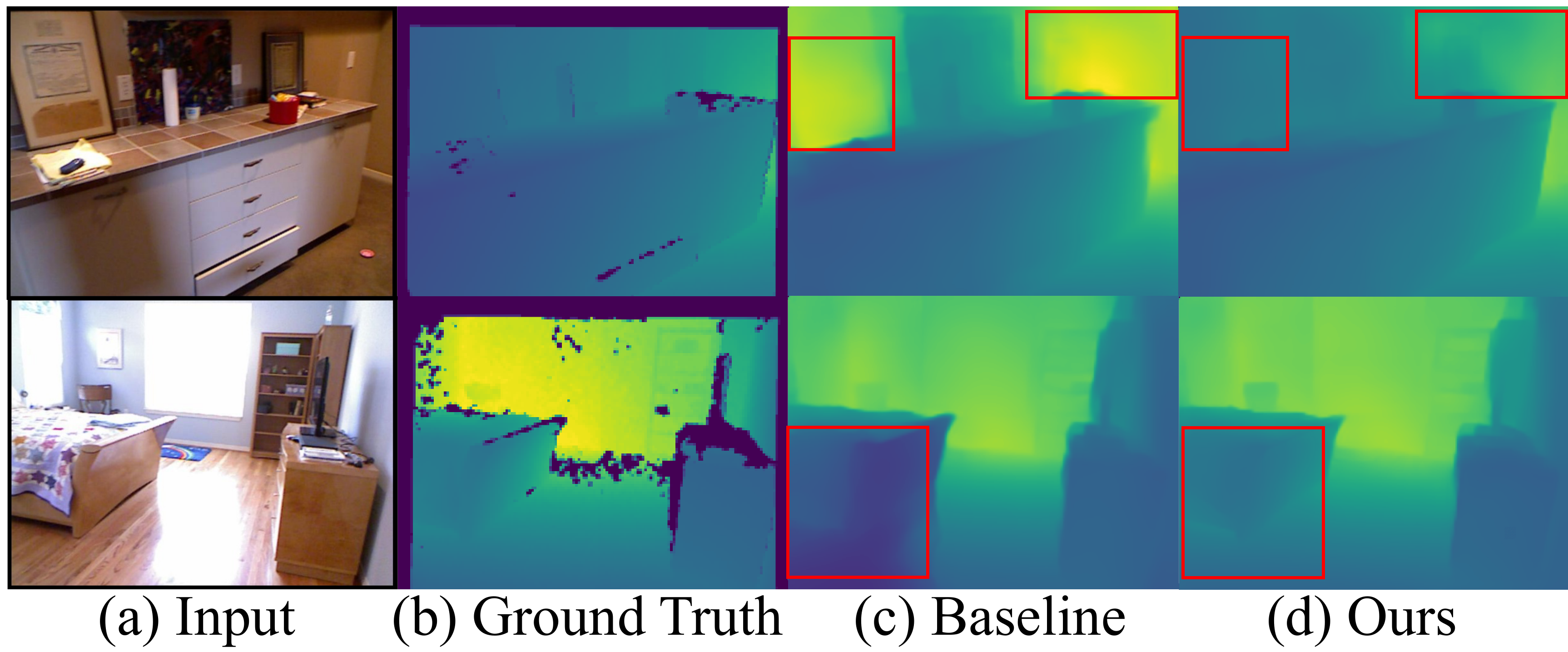}
	\caption{Qualitative comparison of predicted depth maps by our interpretable model and the baseline model~\cite{hu2019revisiting}. The red boxes highlight the difference of the two models.
	}
    \label{fig:pred_visual}
    \vspace{-6mm}
\end{figure}

Fig.~\ref{fig:visual} further shows some example comparisons of feature maps. To demonstrate the effectiveness of our method, pixels are highlighted if its predicted depth is assigned to the corresponding units. We observe that our model has feature maps that are more consistent with pixels of the corresponding depth, showing better interpretability.
From these quantitative and qualitative results, we conclude that our method is able to significantly improve the interpretability of deep networks for MDE.
Meanwhile, across all depth prediction metrics in Table \ref{table:result}, our interpretable models are competitive or even outperform the baseline counterparts, showing that it is possible to enhance the interpretability of an MDE deep network without harming its accuracy.
Fig.~\ref{fig:pred_visual} provides some qualitative comparisons of depth predictions between our interpretable model and the baseline. 

\vspace{-4mm}
\paragraph{Direct Regularizing versus Assigning.}

We quantitatively compare our method of assigning depth to units (cf. Section~\ref{assign}) with the direct approach of regularizing the depth selectivity (c.f Section~\ref{regularize}). As shown in Table~\ref{table:regularizing}, although models trained via directly regularizing the selectivity achieve comparable performance with those trained by our assigning approach, their depth selectivity is much lower due to the issue stated in Section~\ref{regularize}.

\begin{figure*}[!t]
    \centering
	\includegraphics[width=0.83\textwidth]{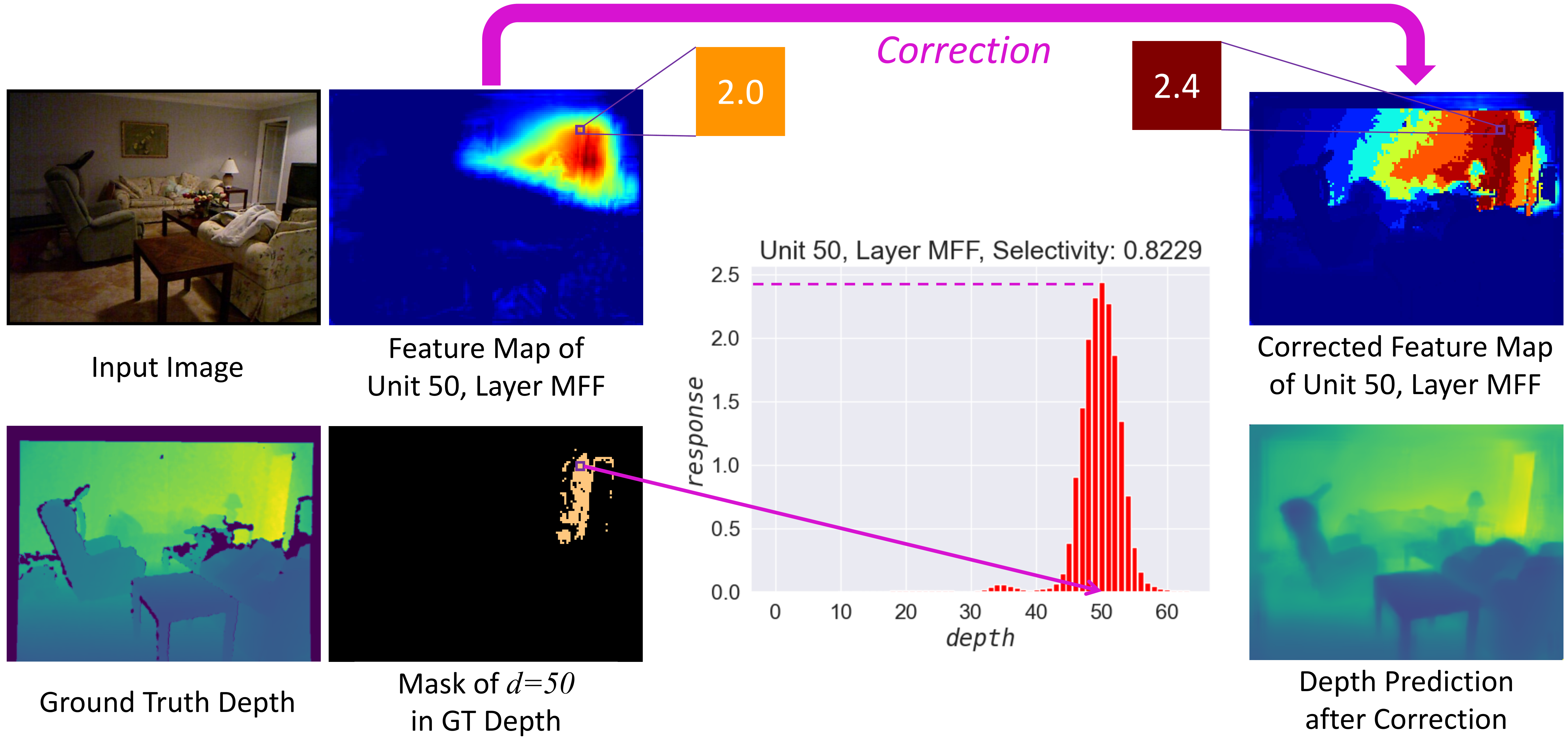}
	\vspace{-1mm}
	\caption{Illustration of our correction operation (cf. Section~\ref{correction}). Suppose a pixel of a unit's feature map has the value of $2.0$, and its corresponding depth ground truth is $50$ after being discretized. From dissection, we know the correct response for this unit on depth $50$ is 2.4, so we can get a new depth prediction map after the response is corrected.
	}
    \label{fig:correction}
    \vspace{-3mm}
\end{figure*}

\subsection{Reliability of Selective Units via Correction}
\label{correction}
\vspace{-2mm}
We further design an experiment to validate the reliability of the selective units. Considering a case where internal units are selective but have zero or little effect to the final output of the model, these interpretable units do not enhance the interpretability of the entire model.
Previous works evaluate the importance of units by ablation, but it is shown that there is only little impact to the accuracy of the model~\cite{DBLP:conf/iclr/MorcosBRB18} when one-by-one ablating each unit. Here, we propose another method to evaluate the reliability of these units, by \textit{correcting} the units instead of ablating them.

Fig.~\ref{fig:correction} illustrates the process of the correction. Here, we define the correct response of a unit for a pixel as its average response in the training data on that pixel's ground truth depth. To be specific, the ground truth depth map is resized to the size of feature maps of a unit using nearest interpolation. Then, for every pixel of the feature map, its value is corrected based on its corresponding ground truth depth and its average response collected from training data. 

\begin{table}[!t]
    \small
	\caption{Performance evaluation before and after correction, where R50 and S154 denote ResNet-50 and SENet-154, respectively (cf. Section~\ref{correction}). In each result, we indicate the performance change from the model without correction to the one after correction using the $\rightarrow$ symbol.}
	\vspace{-2mm}
	\begin{center}
		\begin{tabular}{c|c|c}
		    Model & $\delta_{1.25}$ $\uparrow$ & RMS $\downarrow$ \\
		    \hline\hline
		    \cite{hu2019revisiting} (R50) & 0.849 $\rightarrow$ 0.779 & 0.443 $\rightarrow$ 0.582 \\
		    Interpretable~\cite{hu2019revisiting} (R50) & 0.861 $\rightarrow$ 0.947 & 0.422 $\rightarrow$ 0.362 \\
		    \hline
		    \cite{hu2019revisiting} (S154) & 0.874 $\rightarrow$ 0.856 & 0.409 $\rightarrow$ 0.466 \\
		    Interpretable~\cite{hu2019revisiting} (S154) & 0.882 $\rightarrow$ 0.927 & 0.396 $\rightarrow$ 0.354 \\
		\end{tabular}
	\end{center}
	\label{table:correction}
	\vspace{-6mm}
\end{table}

Table~\ref{table:correction} shows performance evaluation before and after we conduct the correction operation. It is shown that the performance of our models is largely improved after the units' response is corrected, which indicates that units are responsible for the final prediction of the network. Furthermore, our interpretable models demonstrate larger improvement compared to the gain on baseline models using the same correction method.
One reason is that our models are more depth-selective than baseline models, such that the average response contains more information that is related to depth. 
We also note that, the purpose of this evaluation is to validate the reliability of our interpretable units and their effect to the final depth prediction, where the ground truth depth maps are used to achieve such verification but not used in real testing.
\begin{table}[t]
	\renewcommand{\arraystretch}{1.1}
	\setlength{\tabcolsep}{3pt}
	\small
	\caption{Selectivity comparisons on different layers of \cite{hu2019revisiting}.}
	\vspace{-2mm}
	\begin{center}
		\begin{tabular}{c|c|cc|cc}
		     & & \multicolumn{2}{|c|}{Selectivity $\uparrow$ (base)} & \multicolumn{2}{|c}{Selectivity $\uparrow$ (ours)} \\
		    Model & Layer & Training & Testing & Training & Testing \\
		    \hline\hline
		    \multirow{3}{*}{\cite{hu2019revisiting} (R50)} & D\&MFF & 0.4617 & 0.4286 & \textbf{0.8357} & \textbf{0.7529} \\
		     & Rconv0 & 0.4877 & 0.4531 & \textbf{0.7608} & \textbf{0.6846}  \\
		     & Rconv1 & 0.4712 & 0.4399 & \textbf{0.7436} & \textbf{0.6701} \\
		    \hline
		    \multirow{3}{*}{\cite{hu2019revisiting} (S154)} & D\&MFF & 0.4906 & 0.4691 & \textbf{0.8411} & \textbf{0.7693} \\
		     & Rconv0 & 0.5306 & 0.5068 & \textbf{0.7582} & \textbf{0.6945} \\
		     & Rconv1 & 0.4404 & 0.4095 & \textbf{0.7217} & \textbf{0.6626} \\
		\end{tabular}
	\end{center}
	\label{table:sel_layer}
	\vspace{-6mm}
\end{table}

\vspace{-2mm}
\subsection{Applicability of Our Method}
\label{more}
\vspace{-1mm}
\paragraph{More Results on Layers, Models, and Datasets.}

We further apply our method on different layers, models, and datasets to explore its effectiveness. For networks from~\cite{hu2019revisiting}, we consider layers after the first and second convolutional layers in the refine module (referred as layer ``Rconv0'' and ``Rconv1''). Table~\ref{table:sel_layer} and Table~\ref{table:performance_layer} show that for all different layers, our method improves the interpretability (selectivity) over baseline models, while these interpretable models perform competitively in depth estimation accuracy.
We further consider the current state-of-the-art model from~\cite{lee2019big} with the backbone of  DenseNet-161~\cite{huang2017densely} using its four layers, i.e., the layer before the final convolutional layer, the first, second and third upconv layer nearest to the final output (referred as ``iconv1'', ``upconv1'', ``upconv2'', ``upconv3'' following the definition in the supplementary material of the original paper). We also provide experimental results on another commonly-used dataset in outdoor environment, i.e. KITTI~\cite{geiger2013vision}. We show the results of selectivity and depth estimation accuracy in Table~\ref{table:sel_bts} and Table~\ref{table:performance_bts}\footnote{For KITTI, we use an improved set of ground truth depth maps provided by~\cite{uhrig2017sparsity} to train and evaluate both the baseline and our models, so the performance is better than the reported one in the original paper \cite{lee2019big}.}, which validate that our approach is applicable to these various models on another dataset.

\begin{table}[!t]
	\renewcommand{\arraystretch}{1.1}
	\setlength{\tabcolsep}{5pt}
	\small
	\caption{Depth estimation performance of applying our method on different layers of~\cite{hu2019revisiting}. Note that the first row of each model (denoted as ``-'' in Layer) shows the performance of the original baseline model.}
	\vspace{-2mm}
	\begin{center}
		\begin{tabular}{c|c|cccc}
		    Model & Layer & $\delta_{1.25}$ $\uparrow$ & RMS $\downarrow$ & REL $\downarrow$ & log10 $\downarrow$\\
		    \hline\hline
		    \multirow{4}{*}{\cite{hu2019revisiting} (R50)} & - & 0.849  & 0.443 & 0.124 & 0.054 \\
		    \cline{2-6}
		        & D\&MFF & 0.861 & 0.422 & 0.119 & 0.051 \\
                & Rconv0 & 0.860 & 0.423 & 0.119 & 0.051\\
                & Rconv1 & 0.862 & 0.423 & 0.119 & 0.051\\
		    \hline
		    \multirow{4}{*}{\cite{hu2019revisiting} (S154)} & - & 0.874 & 0.409 & 0.111 & 0.049 \\
		    \cline{2-6}
		        & D\&MFF & 0.882 & 0.396 & 0.109 & 0.047 \\
		        & Rconv0 & 0.881 & 0.396 & 0.110 & 0.047 \\
		        & Rconv1 & 0.883 & 0.395 & 0.108 & 0.047 \\
		\end{tabular}
	\end{center}
	\vspace{-8mm}
	\label{table:performance_layer}
\end{table}

\begin{figure*}[!t]
    \centering
	\includegraphics[width=0.95\textwidth]{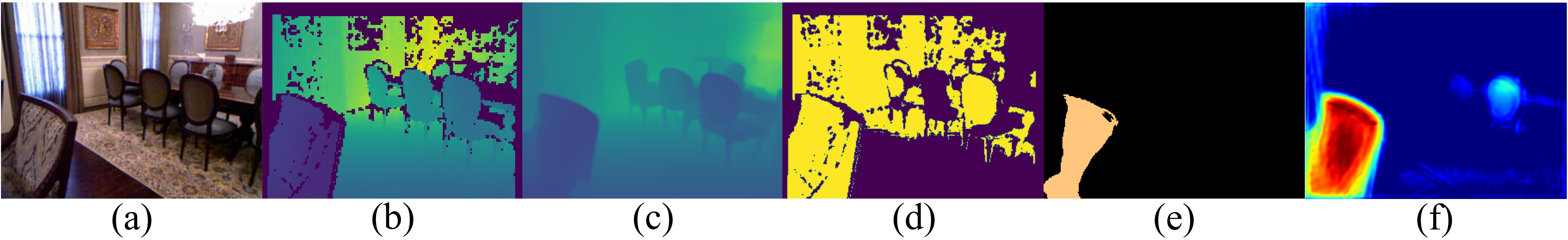}
	\vspace{-2mm}
	\caption{Demonstration of explaining the prediction error on adversarial samples. (a) An adversarial image. (b) Ground truth depth map (c) Predicted depth map. (d) Pixels whose $\delta$ error is above 1.25. (e) Pixels whose predicted depth is within the 11\textsuperscript{th} depth bin. (f) Feature map of Unit 11 in layer MFF.}
    \label{fig:adv}
\end{figure*}

\begin{table*}[!t]
	\renewcommand{\arraystretch}{1.1}
	\setlength{\tabcolsep}{5pt}
	\caption{Depth selectivity and performance of the monocular depth completion model CSPN~\cite{cheng2018depth} and our interpretable counterpart.}
	\begin{center}
		\begin{tabular}{c|cc|cccccc|cc}
		     & \multicolumn{2}{|c|}{Selectivity $\uparrow$} & \multicolumn{6}{|c|}{Depth Accuracy $\uparrow$} & \multicolumn{2}{c}{Depth Error $\downarrow$} \\
		    Model & Training & Testing & $\delta_{1.02}$ & $\delta_{1.05}$ & $\delta_{1.10}$ & $\delta_{1.25}$ & $\delta_{1.25^2}$ & $\delta_{1.25^3}$ & RMS & REL \\
		    \hline\hline
		    CSPN & 0.4022 & 0.4213 & 0.832 & 0.934 & 0.971 & \textbf{0.992} & \textbf{0.999} & \textbf{1.000} & \textbf{0.117} & 0.016 \\
		    Interpretable CSPN & \textbf{0.9394} & \textbf{0.9475} & \textbf{0.860} & \textbf{0.948} & \textbf{0.976} & \textbf{0.992} & 0.998 & 0.999 & 0.119 & \textbf{0.015} \\
		\end{tabular}
	\end{center}
	\vspace{-6mm}
	\label{table:result_cspn}
\end{table*}

\vspace{-4mm}
\paragraph{Application in Depth Completion.}
To show the applicability of our interpretable model, we conduct experiments to apply our method on the monocular depth completion model.
Monocular depth completion is a task highly related to monocular depth estimation, while it additionally takes sparse depth pixels acquired from depth sensors (e.g. LiDAR) or with ground truth depth values as the condition for solving the scale ambiguities and improving the performance of depth estimation.
Here we select CSPN~\cite{cheng2018depth} as our target model. Following their original paper, we adopt the evaluation metrics including accuracy under threshold ($\delta_i < t, t\in \{1.02, 1.05, 1.10, 1.25,1.25^2,1.25^3\}$), RMS, and REL. Table~\ref{table:result_cspn} shows that our approach works well on this model for depth completion, while providing a much better selectivity.

\begin{table}[!t]
	\renewcommand{\arraystretch}{1.1}
	\setlength{\tabcolsep}{2.8pt}
	\small
	\caption{
	Selectivity of applying our method on different layers of~\cite{lee2019big} on the dataset of NYUD-V2~\cite{silberman2012indoor} and KITTI~\cite{geiger2013vision}.}
	\vspace{-2mm}
	\begin{center}
		\begin{tabular}{c|c|cc|cc}
		     & & \multicolumn{2}{|c|}{Selectivity $\uparrow$ (base)} & \multicolumn{2}{|c}{Selectivity $\uparrow$ (ours)} \\
		    Dataset & Layer & Training & Testing & Training & Testing \\
		    \hline\hline
		    \multirow{4}{*}{NYUD-V2} & iconv1 & 0.5202 & 0.3799 & \textbf{0.8456} & \textbf{0.7667} \\
		     & upconv1 & 0.6763 & 0.5507 & \textbf{0.8969} & \textbf{0.8072} \\
		     & upconv2 & 0.5271 & 0.4262 & \textbf{0.8867} & \textbf{0.7929} \\
		     & upconv3 & 0.5476 & 0.4390 & \textbf{0.7871} & \textbf{0.7434} \\
		    \hline
		    \multirow{4}{*}{KITTI} & iconv1 & 0.5000 & 0.4319 & \textbf{0.8321} & \textbf{0.8000}  \\
		     & upconv1 & 0.7128 & 0.5988 & \textbf{0.8935} & \textbf{0.8658} \\
		     & upconv2 & 0.4919 & 0.4181 & \textbf{0.8896} & \textbf{0.8616} \\
		     & upconv3 & 0.5192 & 0.4795 & \textbf{0.8053} & \textbf{0.7893} \\
		\end{tabular}
	\end{center}
	\vspace{-8mm}
	\label{table:sel_bts}
\end{table}

\vspace{-4mm}
\paragraph{Analysis of Model Error.}
Our interpretable model has the advantage of providing a cue to explain why the model makes mistakes. Here we show this application by analyzing the internal representations when predicting on adversarial samples in Fig.~\ref{fig:adv}. We first generate adversarial samples by commonly used white-box attacks FGSM~\cite{goodfellow2014explaining} ($\epsilon=0.05$). As expected, the prediction is not as accurate as before ($\delta_1$ dropped to 0.488 from 0.843). When looking into the mistakes of the prediction, which can be defined as pixels whose $\delta$ error is above 1.25 (\ie, not counted in $\delta_{1.25}$), we find that the predicted depth of a large portion of errors is caused by the 11\textsuperscript{th} depth bin. As we trace back to the neurons, the feature map of Unit 11 shows that the unit is activated on the region with errors, well explaining why those mistakes have been made. The process allows developers and users to know why the model gives unsatisfactory predictions, making the model more trustworthy.

\begin{table}[!t]
	\renewcommand{\arraystretch}{1.1}
	\setlength{\tabcolsep}{5pt}
	\small
	\caption{Depth estimation performance of applying our method on different layers of~\cite{lee2019big} on the dataset of NYUD-V2~\cite{silberman2012indoor} and KITTI~\cite{geiger2013vision}. 
    }
	\vspace{-2mm}
	\begin{center}
		\begin{tabular}{c|c|cccc}
		    Dataset & Layer & $\delta_{1.25}$ $\uparrow$ & RMS $\downarrow$ & REL $\downarrow$ & log10 $\downarrow$ \\
		    \hline\hline
		    \multirow{5}{*}{NYUD-V2} 
		        & - & 0.885  & 0.392 & 0.110 & 0.047 \\
		        \cline{2-6}
		        & iconv1 & 0.882 & 0.389 & 0.110 & 0.047 \\
                & upconv1 & 0.882 & 0.388 & 0.110 & 0.047\\
                & upconv2 & 0.880 & 0.392 & 0.111 & 0.047\\
                & upconv3 & 0.882 & 0.392 & 0.110 & 0.047\\
		    \hline
		    \multirow{5}{*}{KITTI} & - & 0.963 & 2.430 & 0.056 & 0.025 \\
		    \cline{2-6}
		        & iconv1 & 0.961 & 2.435 & 0.059 & 0.026 \\
		        & upconv1 & 0.959 & 2.477 & 0.059 & 0.026 \\
		        & upconv2 & 0.960 & 2.436 & 0.059 & 0.026 \\
		        & upconv3 & 0.960 & 2.415 & 0.058 & 0.026 \\
		\end{tabular}
	\end{center}
	\vspace{-10mm}
	\label{table:performance_bts}
\end{table}

\vspace{-2mm}

\section{Conclusions}
\label{conclusion}

\vspace{-2mm}
In this paper, we propose to investigate the interpretability of deep networks for monocular depth estimation via exploring their internal representations, and advance to make the networks more interpretable. We first find that some hidden units in deep networks for MDE are selective to certain ranges of depth, which inspires us to quantify the interpretability of these networks as the depth selectivity of their internal units.
Furthermore, we propose a simple yet effective method that is applicable to existing deep MDE networks without modifying their original architectures or requiring any additional annotations, showing that it is possible to largely improve the interpretability while not harming or even improving the depth estimation accuracy. Experimental results demonstrate the effectiveness, reliability and applicability of our method. 
In the future work, we will extend our studies on the behavior of internal units in MDE networks to other concepts, such as occlusion boundary, surface normal, and semantics, aiming for a more comprehensive quantification of interpretability and better understanding of deep networks for MDE.

\noindent\textbf{Acknowledgement.}
This paper is supported in part by the National Natural Science Foundation of China under Grant No.61976250 and No.U1811463, in part by the Guangdong Basic and Applied Basic Research Foundation under Grant No.2020B1515020048 and in part by MOST 110-2636-E-009-001, Taiwan. We are also grateful to the National Center for High-performance Computing, Taiwan, for providing computing services and facilities.

{\small
\bibliographystyle{ieee_fullname}
\bibliography{egbib}
}

\end{document}